# Streamlining Systematic Reviews: A Novel Application of Large Language Models

Fouad Trad, ME[1]; Ryan Yammine, MD[2]; Jana Charafeddine, MSc[1]; Marlene Chakhtoura, MD, MSc[2]; Maya Rahme, MSc[2]; Ghada El-Hajj Fuleihan, MD, MPH, FRCP[2]; Ali Chehab, PhD[1]

[1] Department of Electrical and Computer Engineering, American University of Beirut, Beirut, Lebanon

[2] Faculty of Medicine, American University of Beirut, Beirut, Lebanon.

Address queries to:
Fouad Trad, ME
Department of Electrical and Computer Engineering
E-mail: fat10@mail.aub.edu
American University of Beirut, Beirut-Lebanon
P.O. Box: 1107 2020

**Abstract:**

**Background:** Systematic reviews (SRs) are essential to formulate evidence-based guidelines but require time-consuming and costly literature screening. Large Language Models (LLMs) can be a powerful tool to expedite SRs.

**Methods:** We conducted a comparative study to evaluate the performance of a commercial tool, Rayyan, and an in-house LLM-based system in automating the screening of a completed SR on Vitamin D and falls. The SR retrieved 14,439 articles, and Rayyan was trained with 2,000 manually screened articles to categorize the rest as most likely to exclude/include, likely to exclude/include and undecided. We analyzed Rayyan's title/abstract screening performance using different inclusion thresholds. For the LLM, we used prompt engineering for title/abstract screening and Retrieval-Augmented Generation (RAG) for full-text screening. We evaluated performance using article exclusion rate (AER), false negative rate (FNR), specificity, positive predictive value (PPV), and negative predictive value (NPV). Additionally, we compared the time required to complete screening steps of the SR using both approaches against the manual screening method.

**Results:** Using Rayyan, including considered as undecided or likely to include for title/abstract screening resulted in an AER of 72.1% and an FNR of 5%. The total estimated screening time, including manual review of articles flagged by Rayyan, was 54.7 hours. Lowering the Rayyan threshold to 'likely to exclude' reduced the FNR to 0% and the AER to 50.7%, but increased the screening time to 81.3 hours. Using the LLM system, after title/abstract and full-text screening, 78 articles remained for manual review, including all 20 identified by traditional methods. The LLM achieved an AER of 99.5%, specificity of 99.6%, PPV of 25.6%, and NPV of 100%, with a total screening time of 25.5 hours, including manual review of the 78 articles, reducing the manual screening time by 95.5%.

**Conclusions:** The LLM-based system significantly enhances SR efficiency, compared to manual methods and Rayyan while maintaining low FNR.

**Background:**

Systematic Reviews (SRs) are an essential pillar for evidence-based guideline development. However, their process is labor-intensive and time-consuming, requiring authors to screen thousands of articles, with SRs taking on average 67.3 weeks to complete[1]. One of the most time-consuming steps in SRs is literature screening, which is conducted in duplicate and independently, in two steps: title/abstract screening followed by full-text screening.

There is an increasing demand for rapid and frequent SRs by scientists to stay up-to-date in their field as scientific data output is increasing rapidly worldwide, with the corpus of literature doubling every 9 years[2]. Other pressing needs are incurred by pandemics and living practice guidelines.

To meet this increasing demand, many artificial intelligence-based tools have emerged in an attempt to expedite this process, such as Abstrackr, Rayyan AI, ASreviews, Colandr, and DistillerAI[3–7]. These tools vary significantly in their core algorithm and features, but are mostly limited to title/abstract screening. In one review conducted in 2020 comparing various AI-based title/abstract screening tools, Rayyan AI scored the highest in weighted feature analysis[8]. Rayyan AI is a web-based semi-automated screening tool, developed by Qatar Computing Research Institute[4]. It works by feeding words, pairs of words and Medical Subject Headings (MeSH) terms from the titles and abstracts to a Machine Learning (ML) algorithm, more specifically, a Support Vector Machine (SVM) classifier[4].

Recent studies have shown success in the use of Large Language models (LLMs) such as GPT-3.5 and GPT-4 in title/abstract screening[9,10]. Attempts to leverage LLMs for both title/abstract and full text screening are limited[11]. We investigate the use of LLM techniques, such as Prompt Engineering and Retrieval-Augmented Generation (RAG), to automate the aforementioned processes. We propose an end-to-end system powered by GPT-4 that receives an article along with inclusion and exclusion criteria, and then decides whether to include or exclude the article from the SR.

We capitalize on a completed SR on vitamin D and falls to compare the performance of the two ML-based systems, with the traditional manual method as the gold standard [12].

**Methods:**

*Data Preparation*

We used data from a recently completed umbrella review on Vitamin D and Falls[12]. After title/abstract screening, 1,680 full-text papers were reviewed, with 20 SRs of Randomized Controlled Trials (RCTs) included in the final review (Appendix 1).

Manual traditional title/abstract screening followed a validated screening guide (Appendix 2). However, results for 430 articles were inadvertently not saved, reducing the total dataset to 17,346 articles. Importantly, none of the 430 excluded articles were among the final 20 included in the analysis.

The 17,346 articles were imported into Rayyan software. Duplicates were removed using Rayyan's Duplicate Detection Tool.

*Rayyan AI:*

One reviewer trained Rayyan AI by manually screening 2,000 random articles in batches of 100, using the completed umbrella review's title/abstract screening guide (Appendix 2). The reviewer assigned one reason for exclusion for manually excluded articles, following the screening guide.

After each set of 100 articles screened, Rayyan would classify unscreened articles into five categories: "Most Likely To Exclude", "Likely To Exclude", "Undecided", "Likely To Include" or "Most Likely To Include" based on patterns learned during the screening phase. We considered articles rated as "Undecided" or higher to require further manual title/abstract screening, and excluded articles rated as "Likely To Exclude" or lower (Threshold A). Furthermore, we analyzed the results of lowering the threshold for exclusion to "Most Likely To Exclude" only (Threshold B).

After training the model, we evaluated the performance of Rayyan on all unscreened articles including the 20 articles selected in the completed SR. We stopped training Rayyan when the number of unscreened articles in each category stabilized (Figure 1).

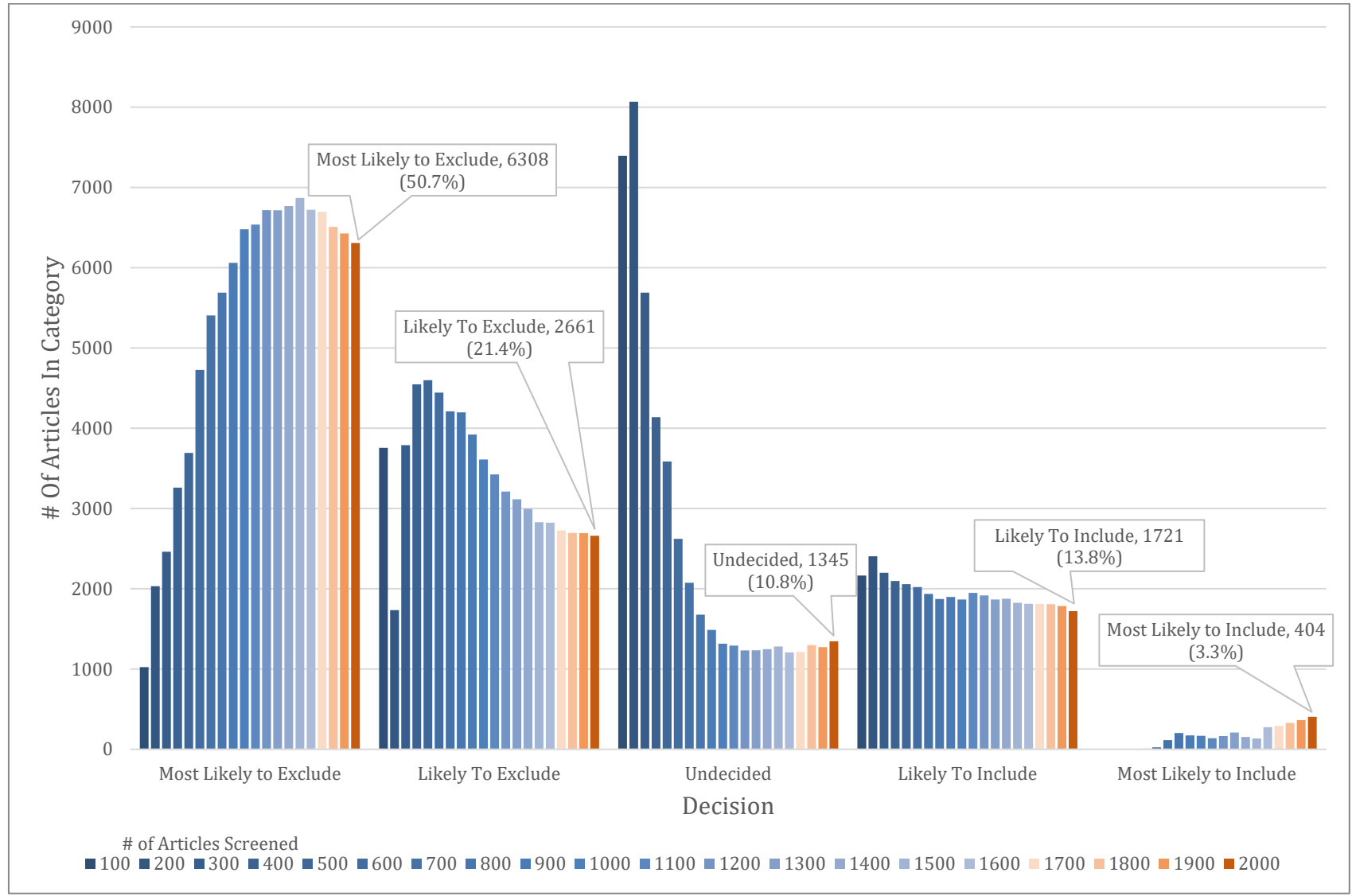

Figure 1: Rayyan classification of unscreened articles into its five default categories with increasing training*

*Training is done in batches of 100, reaching a total of to 2,000 Articles.

***Proposed LLM-based approach***:

The proposed system relies on two LLM techniques: prompt engineering, which involves designing specific input prompts to guide the model's responses, and RAG, which combines external data retrieval with generative capabilities to enhance accuracy and relevance. Together, these techniques automate the two screening phases: Title and Abstract Screening, and Full-Text Screening.

- In title/abstract screening (Phase 1), we input the titles and abstracts of articles into GPT-4 for screening. We give the model a system prompt that instructs it to act as a professional medical researcher performing title/abstract screening. This system prompt helps the model adopt the specified role and respond with the appropriate level of expertise and focus, improving accuracy and consistency during screening (Prompt Engineering). Then, we prompt the model with a series of questions identical to the traditional screening criteria used in the original article's title/abstract screening guide (Appendix 2). The model responds to each question with "yes," "no," or "unsure." When the model is certain about its decision on an article, we proceed accordingly. If the model is uncertain, we retain the article, just as we do with the traditional process, improving sensitivity.
- Articles that pass the first phase undergo a more thorough full-text screening (Phase 2), employing RAG. Here, the full text of each article serves as the document set from which the GPT-4 model retrieves information. A new set of questions identical to the ones used for traditional screening (Appendix 3) is used to evaluate the full texts. The model's responses in this phase to the first five questions are categorized as "yes," "no," or "unsure." Articles are included or excluded similarly to Step 1. The final question prompts the model to identify the outcome studied in the review—falls, fractures, or mortality. The article is only included if "Falls" is one of the outcomes.

The prompts used for both phases can be found in Appendix 4.

To enhance transparency and facilitate an effective review process, the outcomes of all questions are automatically documented in an Excel sheet during both phases for every article (Appendix 5). This logging method enables reviewers to assess the rationale behind the model's decisions. This structured documentation ensures that all decisions are traceable and reviewable. This provides a clear audit trail and supports any necessary re-evaluation of articles automatically screened by the model.

**Statistical analysis:**

We considered the completed SR on Vitamin D and Falls as our gold standard for comparison. For both steps, true positives were defined as articles correctly included for further screening, as they were among the articles included for final analysis, and true negatives were as articles correctly excluded. False positives were articles included by the model for further manual screening but excluded not ultimately included after traditional full-text screening using the manual method, while false negatives were articles excluded by the model but included after manual traditional full-text screening. We used these defined values to calculate the performance metrics described below.

For title/abstract screening using both methods, we evaluated false negative rate (FNR) and article exclusion rate (AER). AER is defined as the total number of automatically excluded articles divided by the total number of articles at the beginning of the relevant step.

For full-text screening, which was assessed using the LLM model only (Rayyan does not support this step), we evaluated FNR, AER, specificity, positive predictive value (PPV), and negative predictive value (NPV). We also assessed these performance metrics from start to end (title/abstract, and full text screen) using the LLM approach.

To estimate workload reduction, we considered both the AER and the time taken to complete screening of the remaining articles. Additionally, FNR was calculated to assess the risk of erroneously excluding relevant articles.

We estimated time required for each screening method as follows:

For the traditional screening method, we estimated the time required for both title/abstract screening (M1) and full-text screening (M2).

For title/abstract screening using Rayyan AI, we calculated the time taken to train the model (R1) and the time needed for manual title/abstract screening of articles remaining after automatic screening (R2). The total time for this process was R = R1 + R2. We estimated M1, R1, and R2 based on the time it took the reviewer to screen 100 articles for Rayyan's training.

For the LLM-based model, we recorded time for automatic title/abstract screening (S1) and full-text screening (S2). Additionally, we estimated time required for manual full-text screening of articles remaining after the automatic process (SM). SM and M2 were calculated based on our team's experience, which estimated that manually screening one full-text article takes an average of 15 minutes

The total time required to complete title/abstract, and full-text screening using the LLM system was S1 + S2 + SM, where SM is equal to 15 minutes multiplied by the number of remaining articles.

**Results:**

**Rayyan title/abstract screening:**

Of the original 17,346 articles, 2,907 articles were deleted after duplicate removal, and 14,439 remained. The reviewer took approximately 1 hour to perform title/abstract screening on 100 articles. Screening all 14,439 articles would take them approximately M1=144.4 hours.

Of the 2,000 articles screened manually to train Rayyan, 1,727 (86.35%) were excluded, and 273 (13.65%) were included. This step took approximately R1=20 hours. Of the remaining 12,439 unscreened articles, Rayyan classified 6,308 (50.7%) as most likely to exclude, 2,661 (21.4%) likely to exclude, 1,345 (10.8%) undecided, 1,721 (13.8%) likely to include and 404 (3.3%) most likely to include (Figure 1). Of the 20 articles included for final analysis in our traditional manual method : 3 were ranked as most likely to include, 6 were ranked as likely to include, 10 were ranked as undecided and 1 was ranked as likely to exclude.

When using Threshold A, 8,969 (72.1%) of the 12,439 unscreened articles were excluded, with the remaining 3,470 (27.9%) articles to be screened manually (Figure 2A), with 1 false negative result. This resulted in: AER 72.1%, FNR 5%, reducing the time needed to complete title/abstract screening using Rayyan to 54.7 hours, an 89.7-hour reduction (62%) when compared to the traditional methods (Figure 3A).

In contrast, when using Threshold B, excluded articles decreased to 6,308 (50.7%), and the remaining 6,131 (49.3%) would undergo further manual screening (Figure 2B), with no false negative results. This inclusion threshold resulted in an AER of 50.7%, and an FNR of 0%, reducing the time needed to complete title/abstract screening using Rayyan to 81.3 hours, a 63.1-hour reduction (44%) when compared to the traditional methods (Figure 3B).

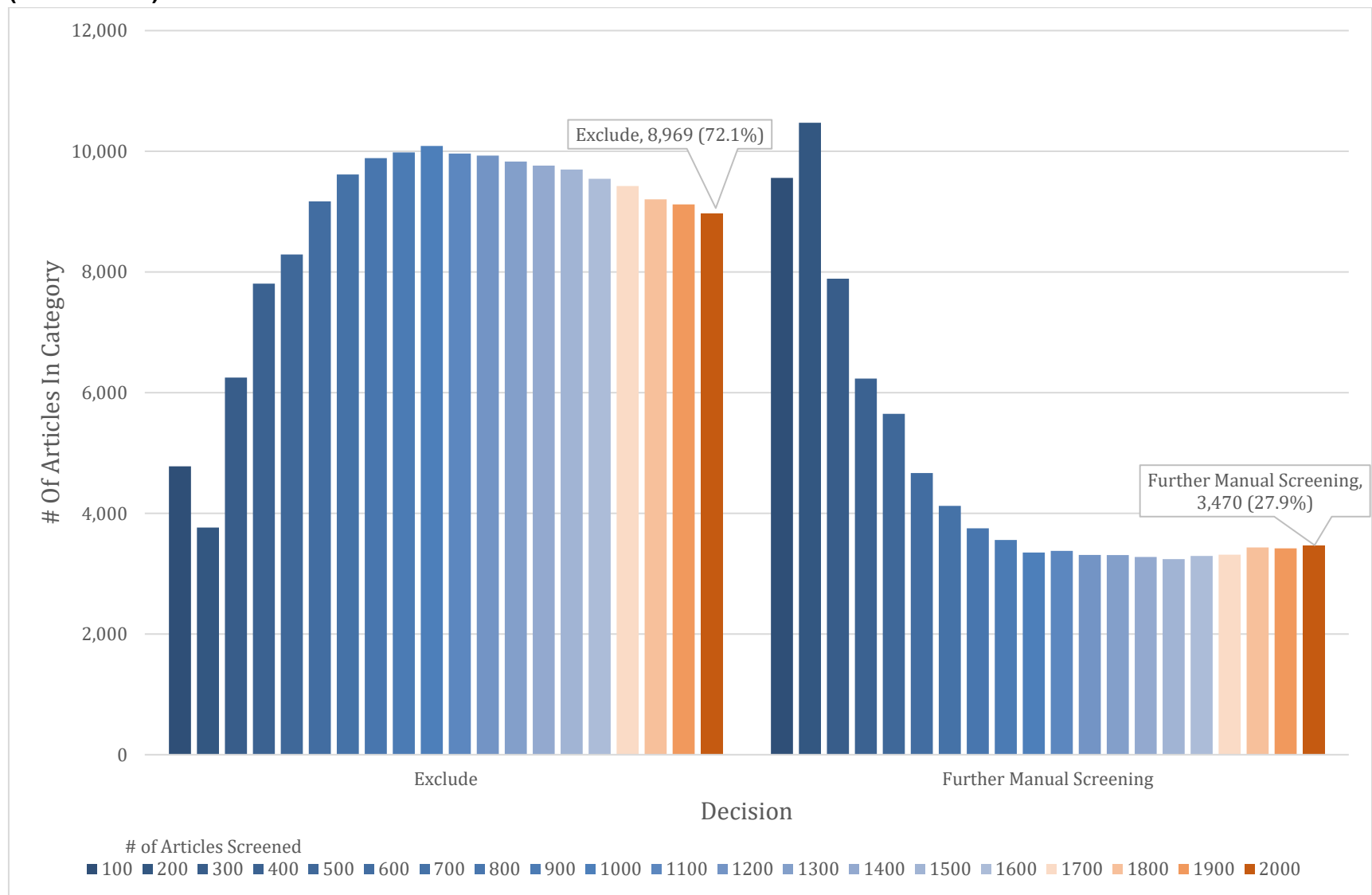

Figure 2A: Rayyan classification of articles as "Excluded" and "Further Manual Screening" with increasing training (Threshold A)*

*"Undecided" is used as threshold for inclusion, training is done in batches of 100, reaching a total of to 2,000 Articles.

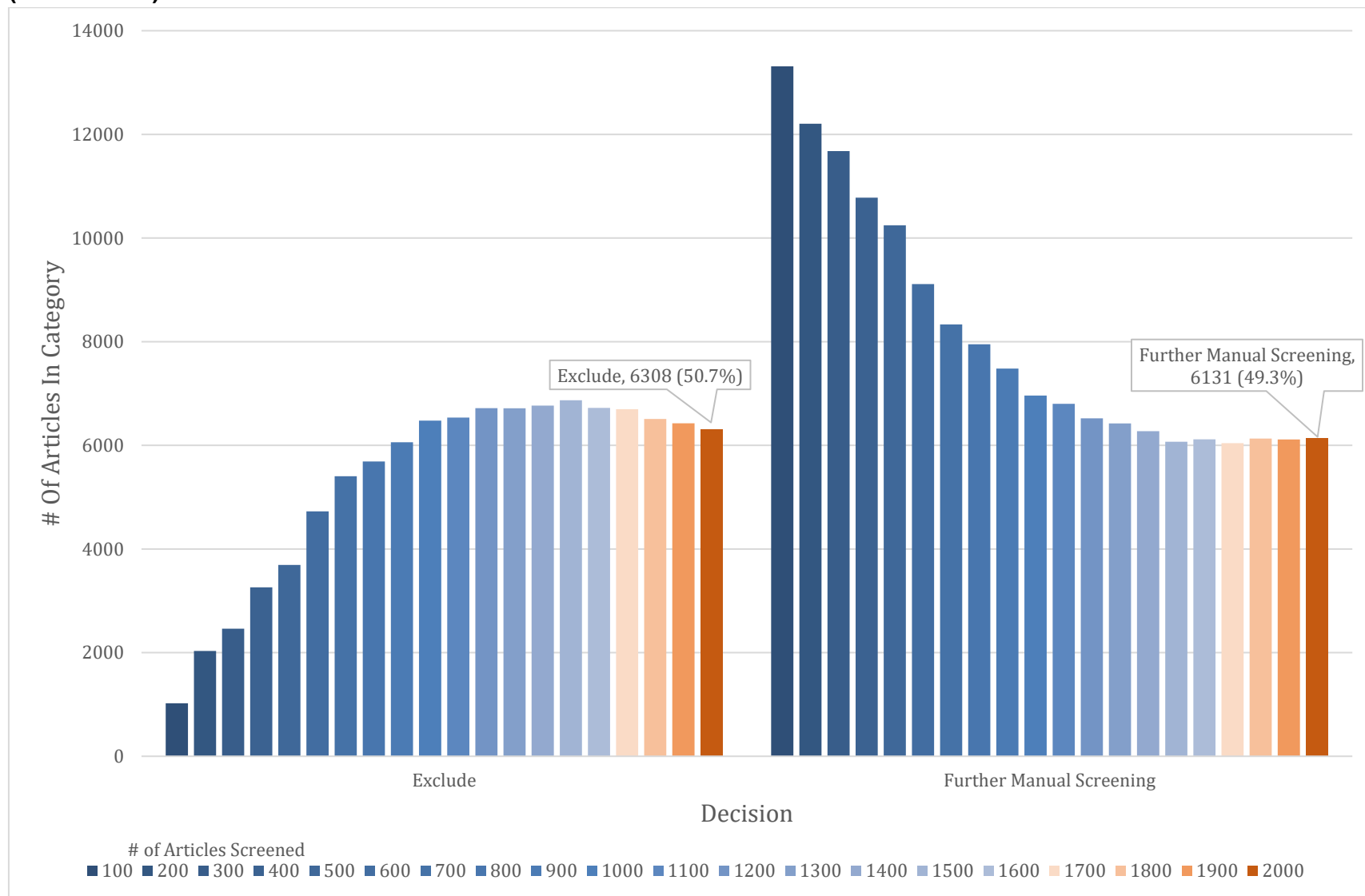

**Figure 2B:** Rayyan classification of articles as "Excluded" and "Further Manual Screening" with increasing training (Threshold B)*

*"Likely to Exclude" is used as threshold for inclusion, training is done in batches of 100, reaching a total of to 2,000 Articles.

**Figure 3A: Flow diagram of Rayyan title/abstract screening steps and results using Threshold A* as inclusion criteria**

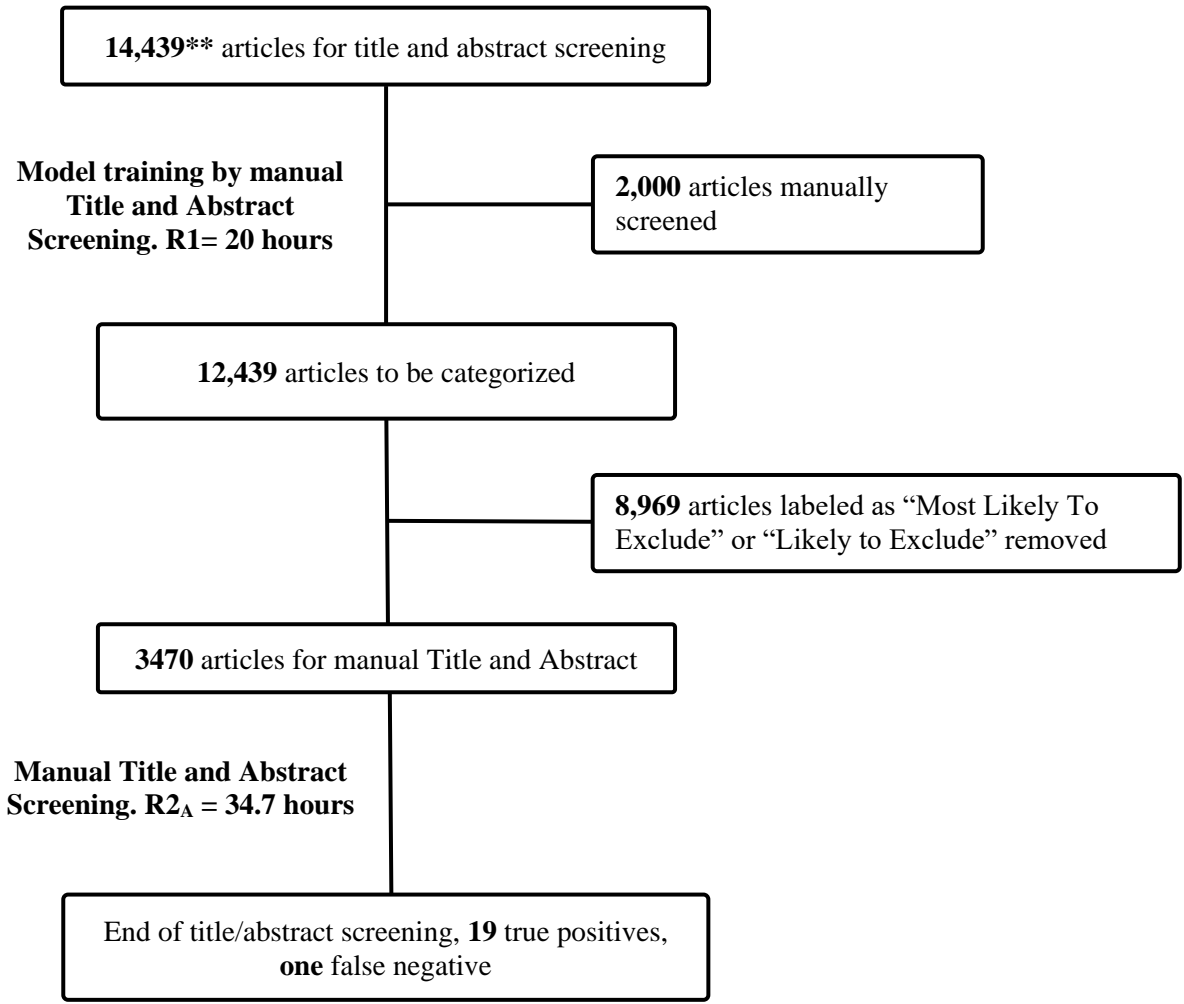

**Total time For Title/Abstract Screening Using Rayyan threshold (A) $R_A$ = R1 + $R2_A$ = 20 + 34.7 = 54.7 hours**

*"Undecided" as threshold for inclusion
**Of the original 17,776 citations, 430 articles were excluded as their results were inadvertently not saved.
2,907 articles were deleted after duplicate removal, of the remaining 17,346 articles and 14,439 remained.

**Figure 3B: Flow diagram of Rayyan title/abstract screening steps and results using Threshold B* as inclusion criteria**

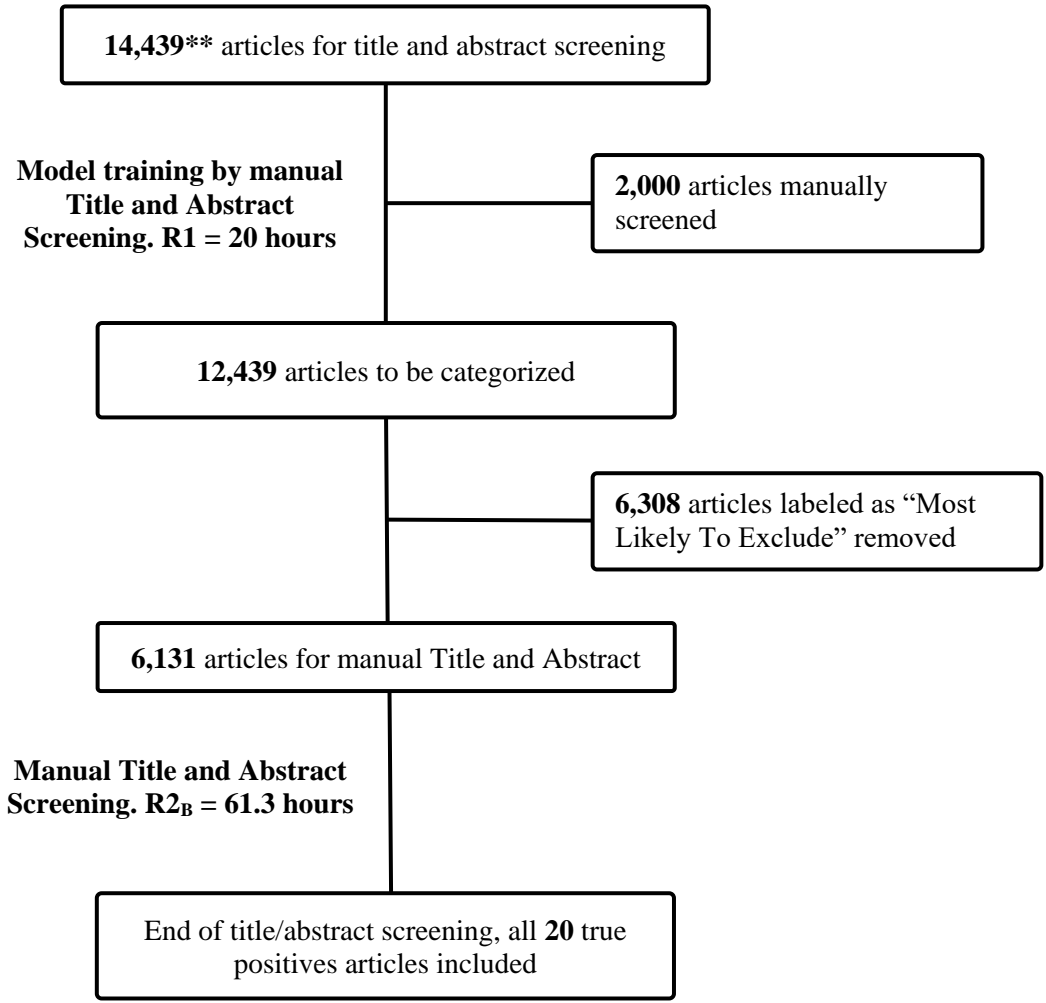

**Total time For Title/Abstract Screening Using Rayyan threshold (B) $R_B = R1 + R2_B = 20 + 61.3 = 81.3$ hours**

*"Likely To Exclude" as threshold for inclusion
**Of the original 17,776 citations, 430 articles were excluded as their results were inadvertently not saved.
2,907 articles were deleted after duplicate removal, of the remaining 17,346 articles and 14,439 remained.

**LLM title/abstract, and full-text screening**

Of the 14,439 articles processed by the GPT-4 model for title/abstract, 3,298 articles (22.8%) met the inclusion criteria and advanced to Phase 2, achieving an AER of 77.2%. None of the 20 retained in the traditional method were excluded, achieving an FNR of 0%. This step took $S_1 = 2$ hours to run.

In the subsequent RAG-based full-text screening phase, the 3,298 full-text articles were evaluated. Out of these, only 78 articles (or 2.37%) were included for manual review, including all 20 articles retained in the traditional method. This step required $S_2 = 4$ hours to run, compared to $M_2 = 1680*15$ minutes = 420 hours for the traditional method. The metrics for this step are as follows: AER: 97.63%, specificity 99.6%, PPV 25.6%, and NPV 100%.

For the entire process, including both phases, the LLM method achieved the following metrics: AER 99.5%, specificity 99.6%, PPV 25.6%, and NPV 100%. Manual screening of the remaining 78 articles would take approximately 19.5 hours (SM), bringing the total time for title/abstract, and full-text screening using the LLM approach to 25.5 hours (Figure 4). This represents a time reduction of 538.9 hours (95.5%) compared to the traditional method, which required an estimated $M_1 + M_2 = 564.4$ hours.

A summary of the performance for both approaches can be found in Tables 1 and 2.

**Figure 4: Flow diagram of LLM Title/abstract and full text screening steps and results**

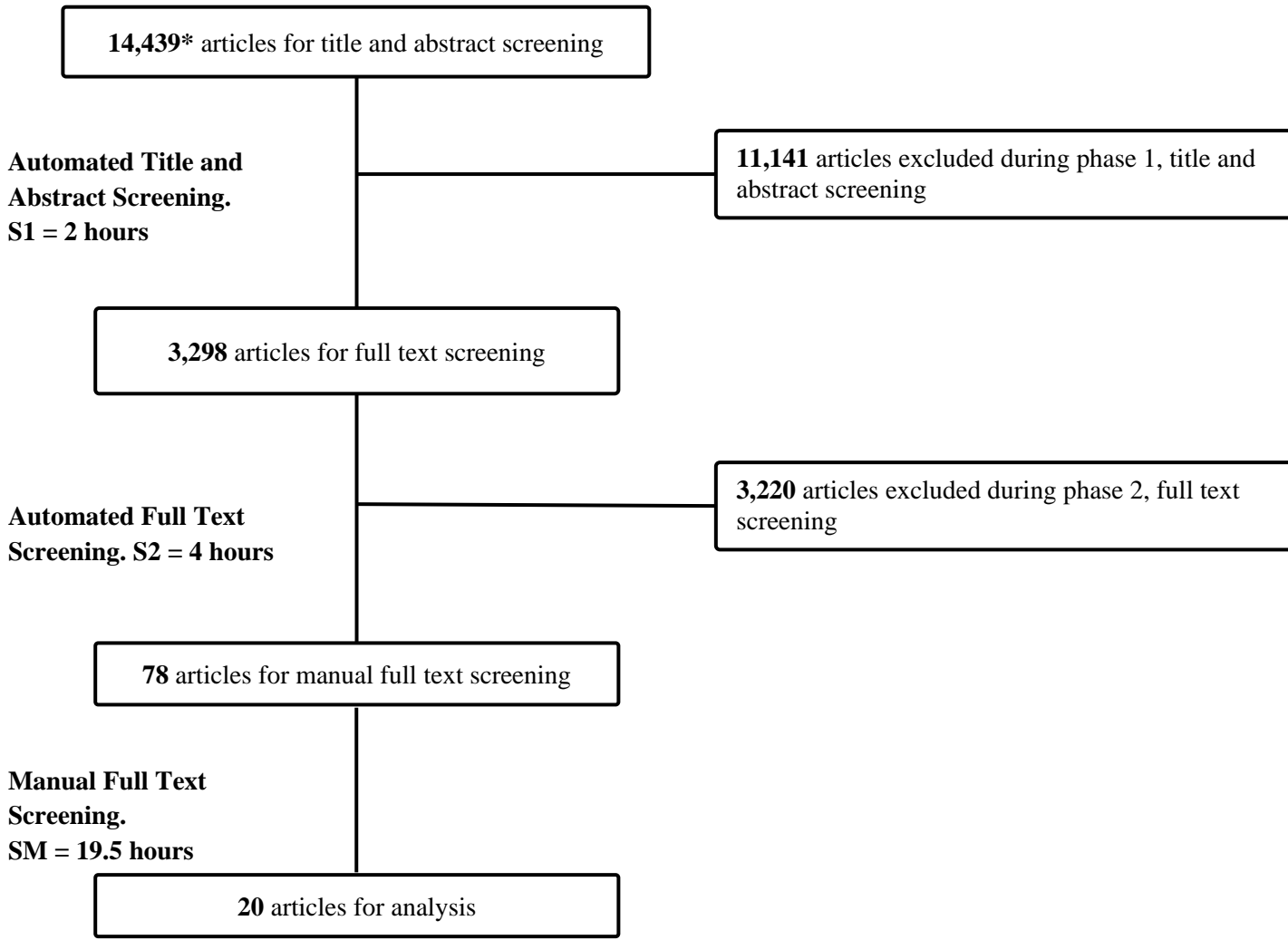

**Total time for Phases 1 and 2 using LLM: S1 + S2 + SM = 2+ 4 + 19.5 = 25.5 hours**

*Of the original 17,776 citations, 430 articles were excluded as their results were inadvertently not saved.
2,907 articles were deleted after duplicate removal, of the remaining 17,346 articles and 14,439 remained.

**Table 1: Summary of Rayyan's performance using both thresholds for title/abstract screening compared to traditional method.**

|  | Rayyan | |
| --- | --- | --- |
|  | **Threshold A** | **Threshold B** |
| **Articles Remaining (AER)** | 3,470 (72.1%) | 6,313 (50.7%) |
| **FNR** | 5% | 0% |
| **R1** | 20 hours | 20 hours |
| **R2** | 34.7 hours | 61.3 hours |
| **Total Time (Title/Abstract)** | 54.7 hours | 81.3 hours |
| **Total Time Saved (Compared to M1 = 144.4 hours)** | 89.7 hours (62%) | 63.1 hours (44%) |

AER: Article Exclusion Rate, FNR: False Negative Rate. R1: time taken to train Rayyan by screening 2,000 articles. R2: time required to complete manual title/abstract screening for articles remained after automated title/abstract screening using Rayyan. M1: Time taken to complete title/abstract screening using traditional methods

**Table 2: Summary of the LLM-based approach's performance for title/abstract and full-text screening compared to traditional method**

|  | LLM | |
| --- | --- | --- |
|  | **Title/Abstract** | **Full-Text** |
| **Articles Remaining (AER)** | 3,298 (77.2%) | 78 (97.6%) |
| **FNR** | 0% | 0% |
| **Time to run** | S1 = 2 hours | S2 = 4 hours |
| **SM** | — | 19.5 hours |
| **Total Time (Title/Abstract+Full-text)** | 25.5 hours | |
| **Total Time Saved (Compared to M1+M2 = 564.4 hours)** | 538.9 hours (95.5%) | |

AER: Article Exclusion Rate, FNR: False Negative Rate. S1: time taken to run the title/abstract screening phase using the LLM approach. S2: time taken to run the full-text screening phase using the LLM approach. M1: Time taken to complete full title/abstract screening using the traditional method. M2: Time taken to complete full-text screening using the traditional method

**Discussion**

Our study shows that both Rayyan AI and the LLM-based system dramatically reduced the workload for SRs compared to traditional methods, while maintaining a low FNR. However, the LLM-based system stood out by not only automating title/abstract screening but also incorporating full-text screening, a more challenging task, through advanced techniques like prompt engineering and RAG. This enabled the LLM to reduce the number of articles for manual full-text review to just 78 out of the original 14,439.

Crucially, the LLM-based system achieved a 95.5% reduction in screening time compared to the traditional method, from 564.4 hours using the traditional approach, to only 25.5 hours. Even more importantly, the LLM maintained a perfect FNR of 0%, meaning no relevant articles were missed during screening. Unlike Rayyan and traditional methods, which rely on human input, the LLM system drastically reduces human intervention, lowering the risks of human error and bias. This impressive combination of time savings and accuracy highlights the LLM's transformative potential for making SRs more efficient and reliable.

According to the Cochrane Collaboration, literature screening ideally involves two reviewers who independently screen articles by following strict screening criteria, to minimize bias, maximize sensitivity, ensuring that no important articles are missed[13]. While few publications have explored the potential of Rayyan software in expediting title/abstract screening, they suffered several drawbacks[14–16]. These include using smaller datasets, with samples varying between 500-1512 articles, and lacking details on thresholds used, rendering an assessment of their performance metrics challenging[16]. Nevertheless, our results based on a larger dataset show a similar high sensitivity using Threshold A[14,15]. However, unlike Valizadeh et al., our study also analyzed a more conservative threshold where false negatives were eliminated[14].

Beyond commercial systems, such as Rayyan, there has been a growing interest in leveraging LLMs to enhance various stages of SRs[17]. Reason et al. evaluated the potential of LLMs to automate tasks such as data extraction, script creation, and report generation within SRs[18]. Others have explored the potential of LLMs in assessment of the quality and risk-of-bias of publications, with varying degrees of success[19–21].

Few publications explored the use GPT-4 to automate title/abstract screening, similar to Phase 1 of our LLM model[10,22–25]. While these studies demonstrated acceptable performance and time savings, they did not extend to full-text screening—a critical and time-consuming phase of SRs. To our knowledge, the only exception is the work of Khraisha et al.[11]. However, their method relied on article segmentation for full-text screening. This approach can affect model performance, as it may struggle to grasp the context when processing segmented parts in isolation, contrary to our RAG framework[11]. As a result, it achieved a low sensitivity of 0.42 and 0.38 during phases 1 and 2, respectively[11]. Importantly, these metrics were based on a limited number of citations screened, 300 titles/abstracts and 150 full texts[11]. In contrast, our system handled a larger dataset of 14,439 articles in the title/abstract screening phase, achieving an FNR of 0% (sensitivity of 100%) during both steps, with a high AER.

None of the discussed publications, including ours, assessed the time needed for the development and finalization of the title/abstract and full text screening sheets. This is an iterative and necessary process with a calibration phase implemented before the sheets are ready for use by any of the three methods. This however does not affect our comparisons between methods. Our study has several strengths. It implemented testing over 14,000 articles to pilot our approach, as opposed to a maximum of 5,634 in other studies also using LLMs[10,22–25]. Additionally, it demonstrated strong performance, with an AER of 99.5%, specificity of 99.6%, PPV of 25.6%, and NPV of 100%, outperforming comparable studies in the literature. Although the LLM-based system requires engineering expertise to build the model, once operational, users can easily interact with it by inputting their inclusion and exclusion criteria in the form of questions. This usability feature underscores the practical application of the system in streamlining the review process. Additionally, the transparent logging of each question's outcome in an Excel sheet not only enhances the system's integrity but also facilitates manual subsequent checks of any article, allowing users to trace decisions back to specific responses, thus reinforcing trust in this approach.

Although the LLM approach offered significant improvements compared to traditional methods and Rayyan, it is still valuable to validate its performance on diverse and complex SRs to confirm its robustness and broader applicability. Future enhancements should focus on refining the logging features to provide even more detailed explanations for each question (knowing why the response was yes, no, or unsure), enhancing explainability and the ability to audit this approach.

**Conclusions**

Our study demonstrates that the proposed LLM-based system significantly enhances the efficiency of the SR process compared to both traditional methods and the commercially available Rayyan system, while maintaining low FNR. Its excellent performance metrics, ease of use, explainability, alignment with traditional methods, and its time efficiency, position it as a very promising approach. Future work could explore expanding the system's capabilities to support more complex review steps, such as data extraction and synthesis.

**List of abbreviations**

SR: Systematic Review

LLM: Large Language Model

RAG: Retrieval-Augmented Generation

PPV: positive predictive value

NPV: negative predictive value NPV

AER: Article Exclusion Rate

FNR: False Negative Rate

RoB: Risk of Bias

MeSH: Medical Subject Headings

SVM: Support Vector machine

RCT: Randomized Controlled Trial

ML: Machine Learning

**Declarations**

1) ***Ethics approval and consent to participate:*** Not Applicable
2) ***Consent for publication:*** Not applicable
3) ***Availability of data and materials:*** The datasets used and/or analysed during the current study are available from the corresponding author on reasonable request.
4) ***Competing interests:*** The authors declare that they have no competing interests
5) ***Funding:*** None
6) ***Author's Contributions:***
   FT, GEHF and AC contributed to the Conceptualization. FT, JC, MC, MR and GEHF contributed to the Formal Analysis. FT, RY and JC contributed to the Investigation. FT,

RY, JC, MC, MR and GEHF contributed to Methodology development. FT, AC and GEHF contributed to Project Administration. FT and JC contributed to Software development. FT contributed to Validation. FT and RY contributed to Visualization. RY, MC, MR and GEHF contributed to Data Curation. FT and RY contributed to Writing of the original draft. FT, RY, MC, GEHF and AC contributed to the Review & Editing of subsequent drafts. GEHF contributed to Funding Acquisition and Resource Provision. GEHF and AC provided Supervision of the project.

7) **Acknowledgements:** Ryan Yammine would like to acknowledge the training received under the Scholars in HeAlth Research Program (SHARP) that was in part supported by the Fogarty International Center and Office of Dietary Supplements of the National Institutes of Health (Award Number D43 TW009118). The content is solely the responsibility of the authors and does not necessarily represent the official views of the National Institutes of Health.

**Appendix 1: Flow diagram of the systematic reviews and meta-analysis included in the original umbrella review on vitamin D and falls**

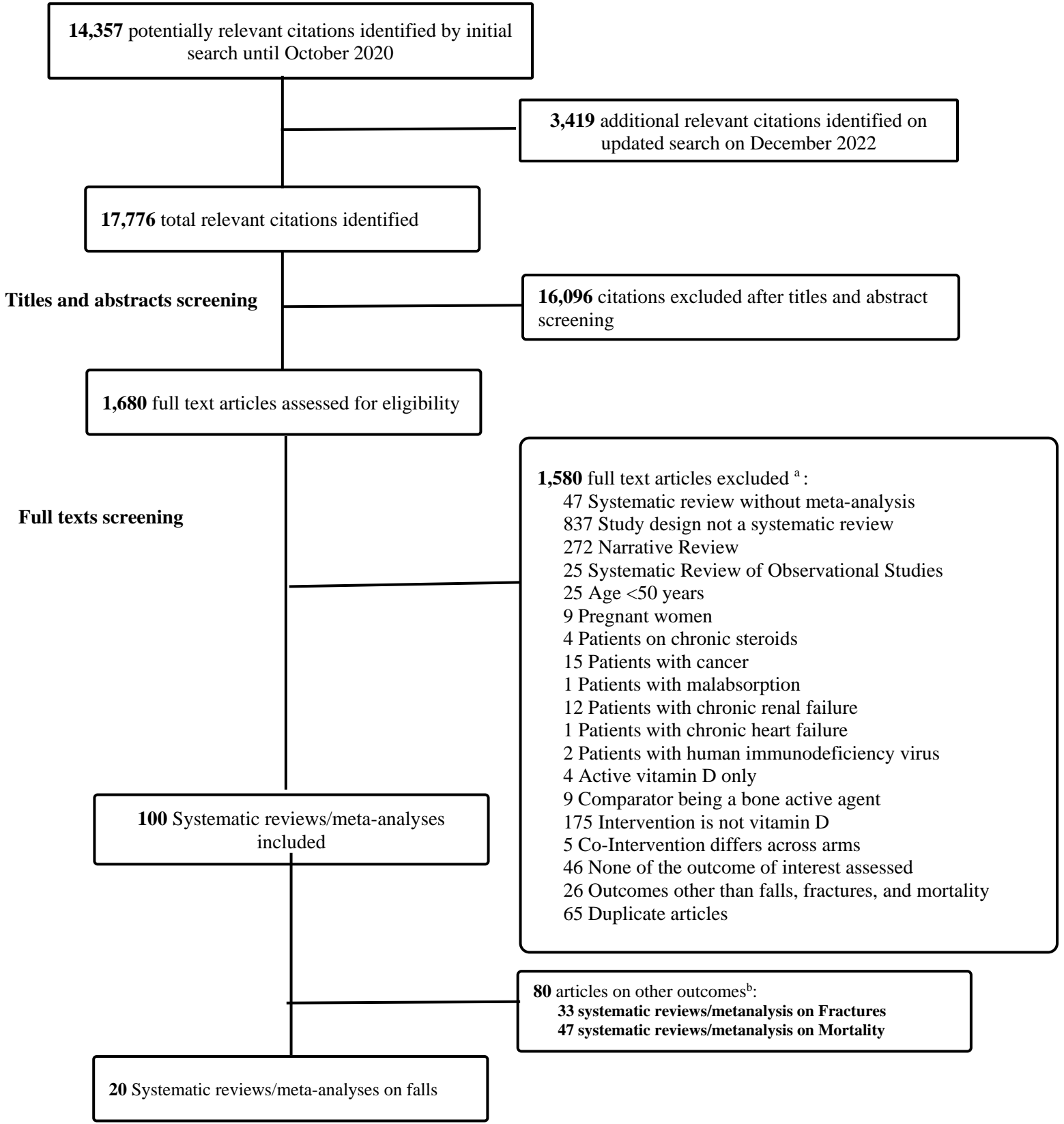

[a] List of excluded articles available upon request. [b] Data on fractures and mortality included in separate manuscripts

**Appendix 2: Title/abstract screening guide**

**Does the study have abstract**
☐ No → Include
☐ Yes → go to the next question

**Is the study design a systematic review and meta-analysis of randomized controlled trials**
☐ No → Exclude
☐ Yes or uncertain → go to the next question
☐ uncertain: narrative review or guideline → Exclude and keep in Sub-Folder Group (Reviews and Guidelines) for later paper

**Is the study population "men and/or women age ≥ 50 years, community dwelling and/or institutionalized"?**
☐ NO → Exclude
☐ Yes or uncertain → go to the next question

**Is the intervention vitamin D, with or without Calcium supplementation?**
☐ NO → Exclude
☐ Yes or uncertain → go to the next question

**Is the comparator vitamin D placebo or control?**
☐ NO → Exclude
☐ Yes or uncertain → Get full text

Exclusion criteria:
-Population: Pregnant women, individuals with advanced diseases (chronic liver failure, chronic renal failure), cancer, individuals on steroids
-Intervention: Active vitamin D: calcitriol, paricalcitol, doxercalciferol, alfacalcidol, falecalcitriol, 22-oxacalcitriol, synthetic vitamin D and vitamin D analogues

**NB: papers on food fortified vitamin D are INCLUDED for now**

**Appendix 3: Full text screening guide**

**Is the study design a systematic review and meta-analysis of randomized controlled trials**
- ☐ No → Exclude (see below)
- ☐ Yes → go to the next question

Exclude if:
- Narrative review (**code 1**)
- Systematic review of observational studies (**code 2**)
- Other study design (**code 3**)

**Is the study population "men and/or women age ≥ 50 years, community dwelling and/or institutionalized"?**
- ☐ NO → Exclude (see below)
- ☐ Yes → go to the next question

Exclude if ≥ 50% of the population:
- Age < 50 (**code 4**)
- Pregnant women (**code 5**)
- Chronic steroids (**code 6**)
- Cancer (**code 7**)
- Patients with malabsorption (**code 8**)
- Patients on anti-fungal (**code 9**)
- Chronic liver failure (**code 10**)
- Chronic renal failure (**code 11**)
- Chronic heart failure (**code 12**)
- COVID-19 (**code 13**)
- Critical Illness (**code 14**)
- ICU (**code 15**)

**Is the intervention vitamin D, with or without Calcium supplementation (± Ca)?**
- ☐ NO → Exclude (see below)
- ☐ Yes → go to the next question

Exclude if:
- Intervention is active vitamin D or vitamin D analogues: calcitriol, paricalcitol, doxercalciferol, alfacalcidol, falecalcitriol, 22-oxacalcitriol, synthetic vitamin D and vitamin D analogues (**code 16**)
- Intervention is not vitamin D (**code 17**)

**Is the comparator vitamin D (± Ca), placebo or control?**
- ☐ NO → Exclude (see below)
- ☐ Yes → go to the next question

Exclude if:
- Comparator is a bone active agent (biphosphonates, Denosumab, HRT, SERM- Raloxifene-, Teriparatide, exercise) (**code 18**)
- If co-interventions (other than Calcium) differ across treatment arms (**code 19**)

**Is one of the outcomes falls, fractures or mortality?**
- ☐ NO → Exclude (**code 20**)
- ☐ Yes → Include

**NB: papers on food fortified vitamin D are INCLUDED for now**

For food (vitamin D) fortified SR, if the co-intervention differs between arms ➔ exclude article.

**Appendix 4 – Prompts used:**

System prompt: Act as an experienced medical researcher performing Title and Abstract screening for Umbrella Review, reviewing articles to decide if they should be included or excluded from your study. Each time you will receive the title and abstract for a paper along with a question that you have to answer based on the authors' work. Later on, your answers will help decide whether an article should be included or excluded so be precise while answering. If you are not sure, answer with 'uncertain'.

Phase 1 - Inclusion questions:
   "Is the study potentially designed as a systematic review and meta-analysis of randomized controlled trials?",
   "Does the study's population include men and/or women age ≥ 50 years, community dwelling and/or institutionalized?",
   "Is there an intervention of vitamin D, with or without Calcium supplementation?",
   "Is the comparator vitamin D placebo or control? Or is this information uncertain? Explain"

Phase 1 - Exclusion questions:
   "Does the paper's population include pregnant women, individuals with advanced diseases like chronic liver failure, chronic renal failure, cancer, or individuals on steroids?",
   "Does the paper primarily focus on active Vitamin D like calcitriol, or paricalcitol, or doxercalciferol, or alfacalcidol, or falecalcitriol, or 22-oxacalcitriol, or synthetic vitamin D or vitamin D analogues?"

Phase 2 questions:
   "Is the study design a systematic review with or without meta-analysis of randomized controlled trials? Answer with 'Yeah', 'Nope' or 'Unsure.",
   "Is the study population's age ≥ 50 years, community dwelling and/or institutionalized"? Answer with 'Yeah', 'Nope' or 'Unsure.",
   "Is the intervention vitamin D, with or without Calcium supplementation (± Ca)? Answer with 'Yeah', 'Nope' or 'Unsure.",
   "Is the comparator vitamin D (± Ca), placebo or control? Answer with 'Yeah', 'Nope' or 'Unsure.",
   "Is one of the outcomes falls, fractures or mortality? Answer with 'Yeah', 'Nope' or 'Unsure.",
   "Is one of the outcomes falls, fractures or mortality? Answer with one of the following categories: 'Falls', 'Fractures', 'Mortality', or 'Unsure'.

**Appendix 5: Screenshot of the sheet used for logging models' answers to the prompts in phase 2.**

| | File name | Article Title | Q1 | Q2 | Q3 | Q4 | Q5 | Outcome Category | Final decision |
|---|---|---|---|---|---|---|---|---|---|
| 39 | 0897190016645328.pdf | Vitamin D and bone health in adults with epilepsy (a systematic review) | Nope | | | | | | excluded |
| 40 | 09637486.2020.1830264.pdf | Comparisons of different vitamin D supplementation for prevention of osteoporotic fractures: a Bayesian network meta-analysis and meta-regression of randomised controlled trials | Yeah | Yeah | Yeah | Yeah | Yeah | Fractures | excluded |
| 41 | 1.pdf | Screening for Vitamin D Deficiency in Adults Updated Evidence Report and Systematic Review for the US Preventive Services Task Force | Yeah | Unsure | Yeah | Yeah | Yeah | Falls, Fractures, Mortality | included |
| 42 | 10.pdf | Association Between Vitamin D Dosing Regimen and Fall Prevention in Long-term Care Seniors | Yeah | Yeah | Yeah | Unsure | Yeah | Falls | included |
| 43 | 11_verylong.pdf | Interventions for preventing falls in older people living in the community | | Yeah | Yeah | Yeah | Yeah | Falls, Fractures | included |
| 44 | 1055-9965.EPI-15-0394.pdf | Serum Retinol and Carotenoid Concentrations and Prostate Cancer Risk: results from the Prostate Cancer Prevention Trial | Nope | | | | | | excluded |
| 45 | 1055-9965.EPI-17-0293.pdf | Serum Insulin, Glucose, Indices of Insulin Resistance, and Risk of Lung Cancer | Nope | | | | | | excluded |
| 46 | 12.pdf | Interventions to Prevent Falls in Older Adults Updated Evidence Report and Systematic Review for the US Preventive Services Task Force | Yeah | Yeah | Yeah | Yeah | Yeah | Falls, Mortality | included |
| 47 | 13.pdf | Interventions to reduce the number of falls among older adults with/without cognitive impairment: an exploratory meta-analysis | Yeah | Yeah | Yeah | Unsure | Yeah | Falls | included |
| | 1351000215Y.0000000018.pdf | Amelioration of persistent left ventricular function impairment through increased plasma ascorbate levels | Nope | | | | | | excluded |